\documentclass{article}
\usepackage{spconf,amsmath,graphicx}
\usepackage[table,xcdraw]{xcolor}
\usepackage{adjustbox}

\title{end-to-end discriminative deep network for liver lesion classification}
\name{\parbox{\linewidth}{\centering Francisco Perdig\'on Romero$^{\star \ddag}$, Andre Diler$^{\star \ddag}$,  Gabriel Bisson-Gregoire$^{\star}$, Simon Turcotte$^{\dagger}$,\\
Real Lapointe $^{\dagger}$, Franck Vandenbroucke-Menu  $^{\dagger}$, An Tang$^{\dagger}$ and Samuel Kadoury$^{\star\dagger}$ \thanks{\ddag \space These authors contributed equally to this work. We  thank Nvidia for GPU donation to our lab and Lisa Di Jorio (Imagia) for support.}}}
\address{$^{\star}$ MedICAL Laboratory, Polytechnique Montreal, Montr\'eal, Canada\\
	$^{\dagger}$ Centre de recherche du CHUM (CRCHUM), Montr\'eal, Canada}

\begin{document}
\ninept
\maketitle
\begin{abstract}

Colorectal liver metastasis is one of most aggressive liver malignancies. While the definition of lesion type based on CT images determines the diagnosis and therapeutic strategy, the discrimination between cancerous and non-cancerous lesions are critical and requires highly skilled expertise, experience and time. In the present work we introduce an end-to-end deep learning approach to assist in the discrimination between liver metastases from colorectal cancer and benign cysts in abdominal CT images of the liver. Our approach incorporates the efficient feature extraction of InceptionV3 combined with residual connections and pre-trained weights from ImageNet. The architecture also includes fully connected classification layers to generate a probabilistic output of lesion type. We use an in-house clinical biobank with 230 liver lesions originating from 63 patients. With an accuracy of 0.96 and a F1-score of 0.92, the results obtained with the proposed approach surpasses state of the art methods. Our work provides the basis for incorporating machine learning tools in specialized radiology software to assist physicians in the early detection and treatment of liver lesions.
\end{abstract}
\section{Introduction}
\label{sec:intro}

\begin{figure}[htb]

\begin{minipage}[b]{1.0\linewidth}
  \centering
  \centerline{\includegraphics[width=7.5cm]{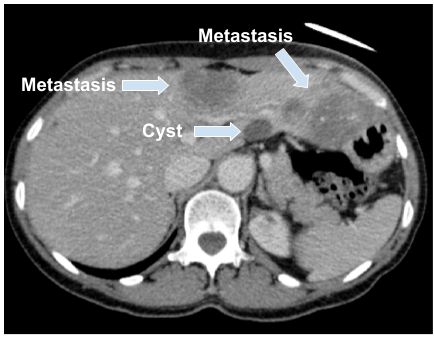}}
\end{minipage}

\caption{Sample axial slice of a CT image, with arrows indicating cysts and metastases with similar contrast levels.}
\label{fig:Axial_Slice}
\end{figure}

Colorectal liver metastasis (CLM) is one of most aggressive liver malignancies, with colorectal cancer (CC) being the second most common cancer in women and third in men with a estimate of 1.36 million cases per year based on the latest report by the WHO \cite{WorldCancerReport}. Due to the difficulty to detect CC in its early stages, there is a very high probability for CLM to develop in almost half of these patients. Computed tomography (CT) is the most commonly used medical imaging modality for the diagnosis and monitoring of patients with liver malignancies. The diagnosis of these malignancies through CT images is a time-consuming task for the specialist and also prone to undetected lesions. On CT images, it is possible to distinguish between liver lesions and healthy tissue because they have different contrasts \cite{WorldCancerReport}. On the other hand, it can be quite challenging for residents or inexperienced specialists to determine whether the lesion is benign, such as in the case of a cyst, or malignant, due to similar contrasts on CT images as illustrated in Figure \ref{fig:Axial_Slice}.

In recent years there has been a growing interest in the development of computer-aided diagnostic tools that help radiologists to classify different types of lesions \cite{SOTA1, SOTA2, SOTA3}. The use of deep learning techniques in medical imaging increased after the success of AlexNet \cite{AlexNet} in the ImageNet competition \cite{ImageNet}.  Deep learning techniques are now not only used for classification, but also for segmentation of different anatomical structures using networks such as U-Net \cite{UNet}. A recent survey presented by G. Litjens et al. \cite{DLSurvey} emphasizes the vast use of deep learning techniques applied to different fields in medical imaging, but also emphasized for the need to properly validate these models with respect to specific medical needs.

In this work, we propose an end-to-end trainable framework for liver lesion classification in CT images using a hierarchical deep network with fully connected classification layers to generate a probabilistic output discriminating between cysts and malignant tumors. The network was trained and validated on a clinical dataset of 63 patients screened for CLM. 

\section{Related Work}
\label{sec:pagestyle}

In recent years, several methods were proposed to accurately classify liver lesions. A. Quatrehomme et al. used a trained classifier with support vector machines (SVM), in which several visual characteristics were extracted using Unser histograms, Law measures, Gaussian Markov random fields, and histogram statistics \cite{SOTA1}. Based on the confusion matrix, the reported accuracy of cysts/metastases classification yielded a rate 90.4\%, which was trained on an in-house dataset that contained 95 CT lesion images. In \cite{SOTA2}, the authors performed a classification of lesions using multidimensional persistent homology, a bottleneck distance, and an SVM classifier. The dataset used for training and validation contained 132 CT lesion images and yielded a classification accuracy of 85\%. Recently, K. Yasaka et al. \cite{SOTA3} used a deep learning classifier of liver lesions in CT images that was trained with 1068 lesion images. The authors tested different models based on convolutional neural networks, yielding an overall accuracy of 84\%. The main limitations with these methods is the lack of a discriminant framework when training the models in order to distinguish cysts from metastases. 

\section{Methods and materials}
\label{sec:format}

\subsection{Dataset}

In this study, a dataset consisting of CT images from 63 consenting patients treated at the Centre hospitalier de l'Universit\'e de Montr\'eal was used, identified from a prospectively maintained hepatobiliary biobank. The dataset contains 230 lesions with an equal distribution of cysts and metastases. The largest cyst lesion was 13.8mL, while the largest metastasis was 534.6mL. On the other hand, the smallest volumes was 0.018mL and 0.272mL, respectively. The average volumes of metastases and cysts was 24.871 mL and 0.791 mL, respectively. The lesions were automatically segmented using a joint liver/liver-lesion segmentation model build from two fully convolutional networks, connected in tandem and trained together end-to-end \cite{vorontsov2018liver}, which yielded a Dice score over 90\% from the 2017 MICCAI LITS challenge. The previous segmentation helps to minimize the size of the image, thus decreasing the computational cost, and also helps the convolutional network from isolating several lesions contained in the same axial slice. For metastases, the primary cancer are variants of malignant neoplasm of rectum or colon.

\subsection{Proposed models}
In this work, we implemented an end-to-end trainable deep network for liver lesion classification. At it's core, we leverage the feature extraction capabilities of two well-known architectures: Inception-V3 \cite{InceptionV3} and InceptionResNet-V2 \cite{InceptionResNet}. Two fully connected layers, each with a Dropout of 0.4 and ReLU activations, were used for the classification. Finally, a fully connected layer of size 2 was added to the architecture, due to the number of classes in the classification problem, followed by a Softmax layer (see Table \ref{tab:Architecture}). 

\begin{table}[htb]

\begin{center}
\caption{Details of the framework architecture.} \label{tab:Architecture}
\begin{tabular}{ccc}
\begin{tabular}[c]{@{}c@{}}Block\\ Type\end{tabular}          & \begin{tabular}[c]{@{}c@{}}Block\\ Name\end{tabular}                      & \begin{tabular}[c]{@{}c@{}}Feature\\ Number\end{tabular} \\ \hline
\rowcolor[HTML]{EFEFEF} 
\begin{tabular}[c]{@{}c@{}}Feature\\ Extraction\end{tabular}  & \begin{tabular}[c]{@{}c@{}}Inception-V3 /\\ InceptionResNet-V2\end{tabular} & 2048                                                     \\
\begin{tabular}[c]{@{}c@{}}Feature \\ Extraction\end{tabular} & Global Average Pooling                                                    & -                                                     \\
\rowcolor[HTML]{EFEFEF} 
Classification                                                & Fully connected                                                                     & 512                                                      \\
Classification                                                & Dropout(0.4)                                                              & -                                                      \\
\rowcolor[HTML]{EFEFEF} 
Classification                                                & Fully connected                                                                     & 512                                                      \\
Classification                                                & Dropout(0.4)                                                              & -                                                      \\
\rowcolor[HTML]{EFEFEF} 
Classification                                                & Fully connected                                                                    & 2                                                       
\end{tabular}
\end{center}
\end{table}

\begin{figure}[htb]
\begin{minipage}[b]{1.0\linewidth}
  \centering
  \centerline{\includegraphics[width=8.5cm]{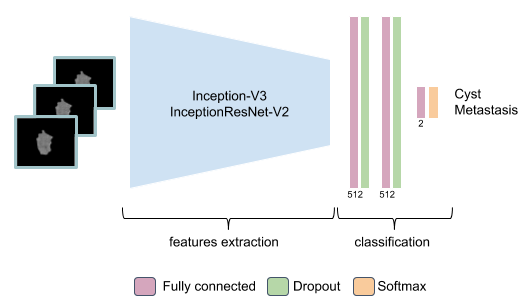}}
\end{minipage}

\caption{Framework architecture.}
\label{fig:framework_arch}
\end{figure}

The inputs to the network were composed of 3 orthogonal patches, where the first image belongs to the plane where the lesion is larger, while the other two images belong to the adjacent orthogonal slices. This ensures that the network is provided enough information for optimal classification accuracy (see Figure \ref{fig:framework_arch}).

The proposed model is based on the Inception-V3 \cite{InceptionV3} architecture, following improvements from the original model \cite{InceptionV1}. The architecture addresses the common problem of high variations in the spatial location of targets. In medical images, the same object can be observed at different scales. Choosing the appropriate kernel size for convolutions is challenging as  tumors and cysts can greatly vary in size (Figure \ref{fig:Axial_Slice}). Moreover, the computational cost increases with the kernel size. To solve this problem, a concatenation of multiple feature extraction methods were used. Extraction with different kernel sizes and pooling steps are processed in parallel, and all extracted features are concatenated at the end of the Inception module. The best feature extraction filters are chosen through back-propagation. 

The architecture implements the concept of factorization by asymmetric convolutions, which is a simple but efficient way to optimize matrix operations. This is done by reducing nxn convolutions into nx1 followed by 1xn convolutions, improving improved compared to the previous factorization of convolutions and reduced computational cost. For example, a pair of 3x3 convolutions can be replaced by one 5x5 convolution (reduction factor of 2.78). Although, a representational bottleneck is created, this does not affect performance given a properly balanced size, based on the hypothesis of a correlation presence between adjacent pixels in the image. This correlation makes the reduction of the image dimension (bottleneck) possible without losing relevant information. To counterbalance the effect of the representational bottlenecks, the filter banks were expanded. Hence, the module was not made deeper but wider, as the depth reduces the dimensions (and the information) and the width preserves it. Instead of putting 3x3 convolutions followed by 3x1 and 1x3 convolutions serially, the output of the 3x3 convolution is fed in parallel to the 1x3 and 3x1 convolutions.

Auxiliary classifiers (AC) were added to improve convergence at the end of the model. They reduce the vanishing gradient problem, by acting as regularizers. AC work well when combined with other regulators like batch normalization and dropout layers. Finally, additional techniques such as batch normalization and label smoothing were added to improve the computationally efficiency of the final architecture without affecting its performance.

In our model, we leverage the use of residual connections present in the InceptionResNet \cite{InceptionResNet} architecture, as they generally improve classification accuracy. The original InceptionResNet combined the Inception modules with residual connections. The so-called residual connections add features that were the input (officially called Identity connection or Identity mapping) of two or three previous convolutional layers. This approach introduced by He in al. \cite{ResNet_ref} demonstrated that the residual connections reduce the time of convergence for a classification network (ILSVRC 2015) as it minimizes the effect of vanishing gradients, especially in very deep networks.

Given the large size of the Inception-V3 architecture, it was expected that the natural evolution of the architecture would be the use of residual blocks. This resulted in reduced convergence time and slightly better results when compared to Inception-V3. Two architectures were proposed: InceptionResNet-V1, which has a computational cost similar to Inception-V3, and InceptionResNet-V2, a more computationally expensive version but with a much higher performance.

\subsection{Training Protocol}

In this section, we present the training protocol for the models, along with parameter selection strategies.

\textbf{Cropping and padding}: All volumes were cropped to the size of the bounding boxes generated by the segmentations during  data preparation. To have the same size for all images, a padding was applied to fit the largest lesion, with a size of 252 x 210 pixels. The value of the padding was the mean value of the extracted bounding box. Using this technique, the mean value of the bounding box image is less affected compared to zero padding.

\textbf{Normalization}: Mean centering and standard deviation normalization was applied to each image patch.

\textbf{Train/Validation/Test split}: The main dataset containing 230 lesion images was divided into three sets: 138 lesions images was used as a training set, 46 images as a  validation set, and 46 images as a test set. The images were randomly selected for each separate set. In our experiments, we tested the automatic initialization of the weights and using pre-trained weights in the ImageNet database \cite{ImageNet}.

\textbf{Data augmentation}: Due to the dataset's limited size, data augmentation was used to avoid over-fitting. The data augmentation strategy included rotation (up to 30 degrees), horizontal and vertical shift (up to 25 pixels), and flipping. The previous parameter values were randomly chosen at every epoch, this was done automatically by the deep learning used framework (Keras/Tensorflow).

\textbf{Batch size}: A small batch size of 8 images was used to avoid CPU bottlenecks during training since image batches are generated in real-time on CPU due to the data augmentation strategies.

\textbf{Optimization and Learning Rate Scheduling}: The Adam optimization algorithm \cite{Adam} was used with an initial learning rate of 10$^{-3}$. This learning rate was reduced by a factor of 2 after 10 epochs without improvements in the validation set accuracy, repeatedly. The allowed minimum learning rate was 10$^{-10}$.

\textbf{Iterations}: The initial number of epochs was 10$^{3}$, but due to the early stopping criterion that was used, the final number of epochs was 55. For early stopping, the validation set accuracy was monitored, after 50 epoch without improvements the train is ended.

Training time on an NVIDIA Titan Xp GPU was 170 minutes, while the inference time per single lesion was 3 ms (89 ms on an I7 7th generation CPU). The above timings were calculated for the proposed model, including InceptionNetV3.

\section{Results and Discussion}

Table \ref{tab:Perfomance} presents the classification results in comparison to  a texture-based approach trained with SVMs, InceptionResNetV2 and Inception-V3. It shows the proposed model with pre-trained weights from the ImageNet \cite{ImageNet} performed better in 4 out of the 7 metrics. Despite the fact that the pre-trained weights are obtained from natural images from the ImageNet dataset, using these weights as initialization for the network produced better results due to the important variability between the existing classes in ImageNet ($>$ 1000). Furthermore, due the large number of images provided to the first layers, this allows the model to train the primary filters with highly discriminant features, allowing to distinguish cysts from metastases. This feature makes a model that uses these weights to achieve better accuracy than a model trained from scratch.


\begin{table*}[htb]

\begin{flushleft}
\caption{Classification performance of the trained models. 
IPW: with ImageNet pretrained weights
NPW: no pretrained weights.} \label{tab:Perfomance}

\end{flushleft}

\begin{center}
\begin{tabular}{l|ccccccc}
\textbf{Model}               & \textbf{Accuracy} & \begin{tabular}[c]{@{}c@{}}\textbf{Balanced} \\ \textbf{Accuracy}\end{tabular} & \textbf{F1 score} & \textbf{AUC}   & \textbf{Precision} & \textbf{Recall} & \textbf{Specificity} \\ \hline
SVM with texture features  & 0.73     & 0.71                                                        & 0.63     & 0.78 & 0.70       & 0.81    & 0.65        \\
InceptionResNet-V2 (NPW)  & 0.87     & 0.85                                                        & 0.78     & 0.92 & 0.90       & 0.9    & 0.78        \\
InceptionResNet-V2 (IPW) & 0.89     & 0.90                                                        & 0.83     & \textbf{0.97} & 0.96      & 0.87   & 0.92        \\
Inception-V3 (NPW)     & 0.86     & 0.91                                                        & 0.82     & \textbf{0.97} & \textbf{1.00}         & 0.81   & \textbf{1.00}           \\
\textbf{Proposed model (IPW)}    & \textbf{0.96}     & \textbf{0.93}                                                        & \textbf{0.92}     & \textbf{0.97} & \textbf{1.00}      & \textbf{0.94}      & 0.85       
\end{tabular}  

\end{center}
\end{table*}

\begin{figure}[htb]
\begin{minipage}[b]{1.0\linewidth}
  \centering
  \centerline{\includegraphics[width=8.5cm]{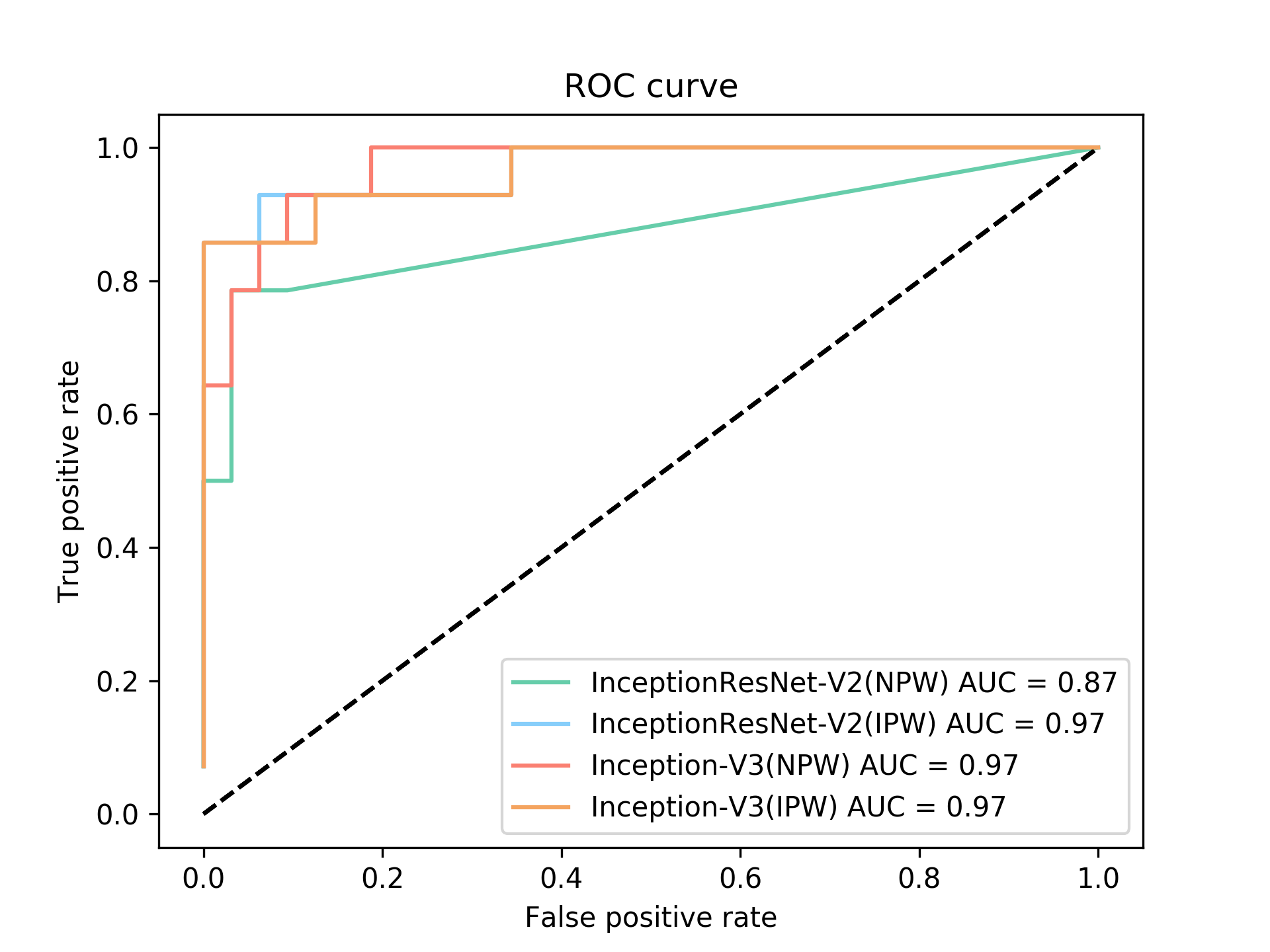}}
\end{minipage}

\caption{Comparison in model performances with ROC curves.}
\label{fig:ROC}

\begin{minipage}[b]{1.0\linewidth}
  \centering
  \centerline{\includegraphics[width=8.5cm]{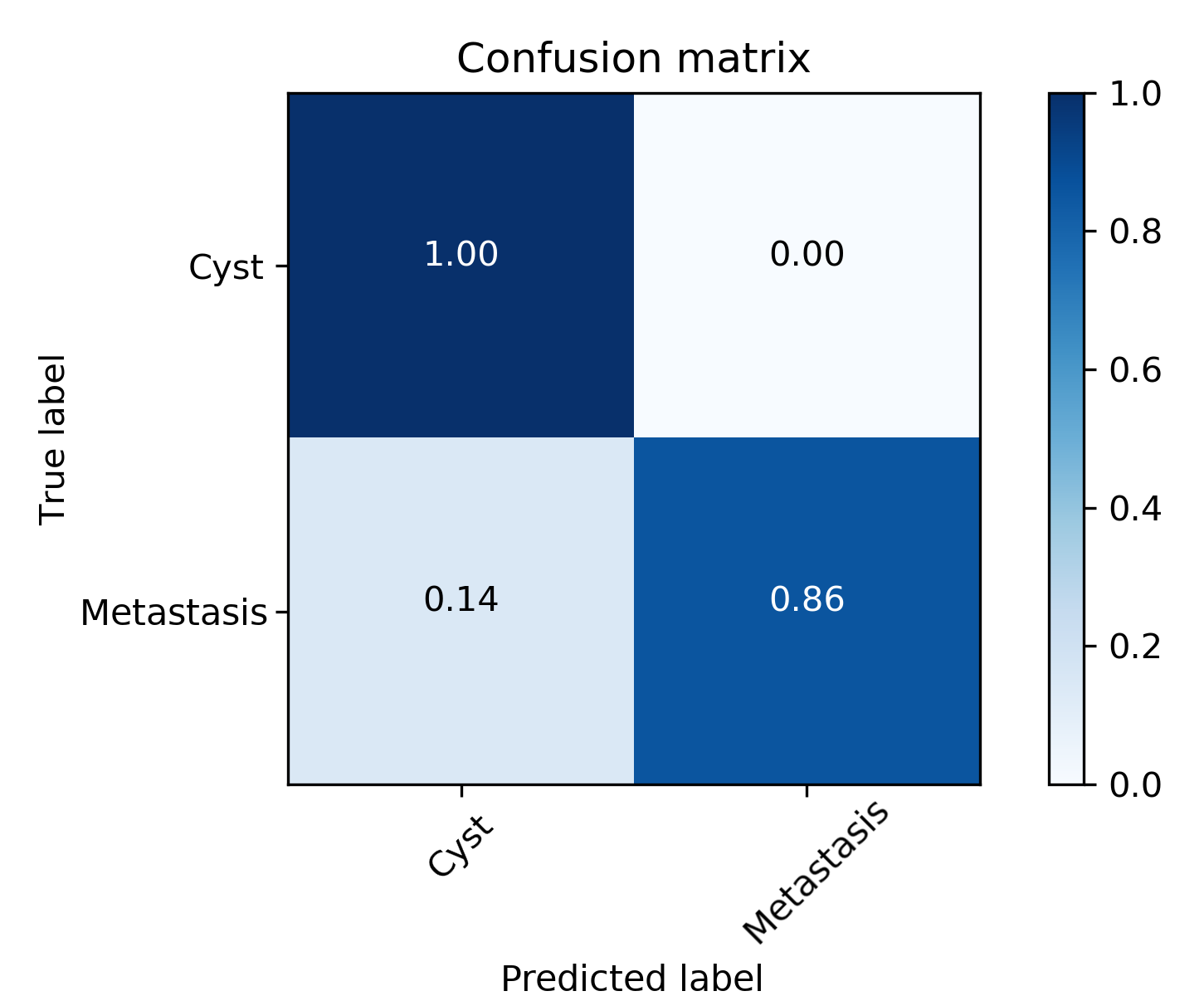}}
\end{minipage}

\caption{
Confusion matrix showing the performance of proposed model based Inception-V3  using pre-trained weights on our dataset.}
\label{fig:CM}
\end{figure}

Figure \ref{fig:ROC} shows the efficiency of the classifiers implemented through the receiver operating characteristic (ROC) curves. Figure \ref{fig:CM} shows the performance of the model with the proposed model based on InceptionV3 using pre-trained ImageNet weights through the confusion matrix. 

\begin{figure}[htb]

\begin{minipage}[b]{1.0\linewidth}
  \centering
  \centerline{\includegraphics[width=7.5cm]{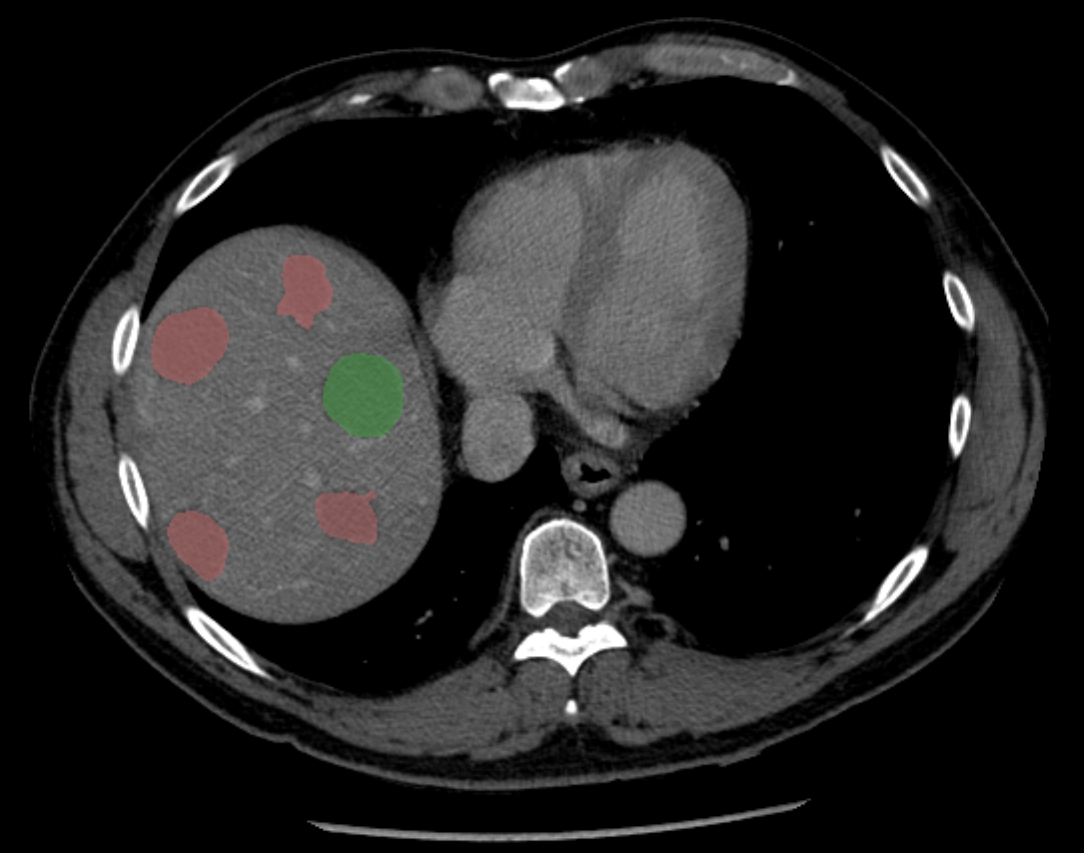}}
\end{minipage}

\caption{Sample classification result; green: cyst, red: metastasis.
}
\label{fig:Cyst_Meta}
\end{figure}

Once lesions are classified with the proposed framework, the different segmented lesions in the CT volume are color-coded, thus allowing specialists to quickly validate the lesion type (see Figure \ref{fig:Cyst_Meta}). Due to the limited number of openly available liver lesion datasets with confirmed presence of malignant tumors, direct comparison with other published results is difficult. However the comparative results reported in this study show the improvement with regards to state of the art methods, including texture-based classifiers. 

\section{Conclusion}
\label{sec:typestyle}

In this work, we proposed a framework to classify cysts and metastases in CT images using deep neural networks from automated segmentations. A  convolutional neural network derived from Inception-V3 was developed, which was trained from scratch using pre-trained weights in the ImageNet dataset and includes fully connected classification layers generating probabilistic outputs. The proposed model yielded the highest accuracy and AUC compared to previous state-of-the-art approaches. The framework can therefore provide a screening tool for early detection of malignant lesions which can be integrated in a CAD software.

\bibliographystyle{IEEEbib}
\bibliography{refs}

\end{document}